\documentclass[letterpaper, 10 pt, conference]{ieeeconf}  

\IEEEoverridecommandlockouts                              

\usepackage[noadjust]{cite}
\usepackage{graphicx}
\usepackage{subfig}
\usepackage{amsmath}
\usepackage{amssymb}
\usepackage{multirow}
\usepackage{multicol}
\usepackage{balance}
\usepackage{csquotes}
\usepackage{hyperref}
\usepackage{tikz}
\usepackage{booktabs}
\usepackage{comment}
\usepackage{stfloats}

\usepackage{float}

\newcommand{\para}[1]{\vspace{0.4em}\noindent\textbf{#1} }
\newcommand{\human}{\textrm{H}}
\newcommand{\robot}{\textrm{R}}
\newcommand{\DKL}{\mathbb{D}_{\mathrm{KL}}}

\DeclareMathOperator*{\argmin}{arg\,min}

 \title{\LARGE \bf Latent Emission-Augmented Perspective-Taking (LEAPT)\\for Human-Robot Interaction}

\author{Kaiqi Chen$^{1}$, Jing Yu Lim$^{1}$, Kingsley Kuan$^{1}$, and Harold Soh$^{1,2}$
\thanks{This research is supported by the National Research Foundation Singapore and DSO National Laboratories under the AI Singapore Programme (AISG Award No: AISG2-RP-2020-016).}
\thanks{$^{1}$Authors are with the Dept. of Computer Science, National University of Singapore. 
        {\texttt{\{kaiqi, jy\_lim, kingsley, harold\}@comp.nus.edu.sg}}
\thanks{$^{2}$Smart Systems Institute, National University of Singapore}
}}

\begin{document}

\maketitle
\thispagestyle{empty}
\pagestyle{empty}

\begin{abstract}
Perspective-taking is the ability to perceive or understand a situation or concept from another individual's point of view, and is crucial in daily human interactions. Enabling robots to perform perspective-taking  remains an unsolved problem; existing approaches that use deterministic or handcrafted methods are unable to accurately account for uncertainty in partially-observable settings. This work proposes to address this limitation via a deep world model that enables a robot to perform both perception and conceptual perspective taking, i.e., the robot is able to infer what a human sees and believes. The key innovation is a decomposed multi-modal latent state space model able to generate and augment fictitious observations/emissions. Optimizing the ELBO that arises from this probabilistic graphical model enables the learning of uncertainty in latent space, which facilitates uncertainty estimation from high-dimensional observations. We tasked our model to predict human observations and beliefs on three partially-observable HRI tasks. Experiments show that our method significantly outperforms existing baselines and is able to infer visual observations available to other agent and their internal beliefs. 
\end{abstract}
\section{INTRODUCTION}
\label{sec:intro}

\begin{figure} [!t]
    \centering
    \vspace{0.5cm}
    \includegraphics[width=1.0\columnwidth]{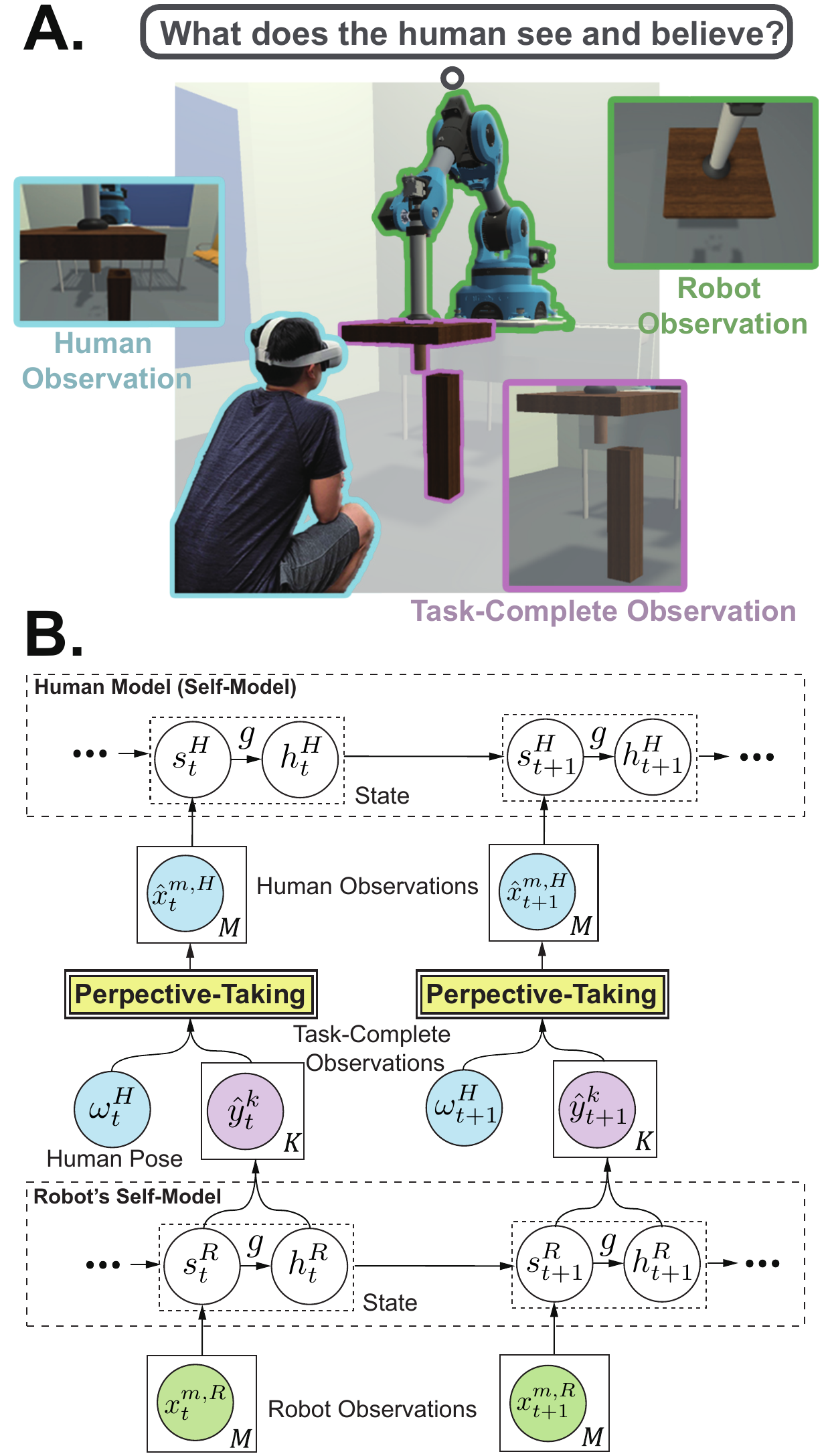}
    \caption{\small Perspective-Taking Example. (\textbf{A}) As the robot (green) is holding the table, it cannot see the peg and hole (purple), which is clearly visible to the human (light blue). The robot considers the perspective of the human and reasons about what the human can observe, despite not knowing what he actually sees. (\textbf{B}) Human belief inference. In brief, the robot's world model (self-model) is used to sample possible environment states and predict human observations, and the human model is used to infer human belief. By leveraging the inference network and perspective taking-model, we can infer human belief under each sampled environment state.}
    \label{fig:moti-example}
\end{figure}

This work focuses on the problem of perspective taking, which is the ability to take another agent's point of view, either visually or conceptually. For example, consider the scenario in Fig. \ref{fig:moti-example} where a robot and a human are aligning a peg and hole to assemble a table. As the robot is holding the table, it cannot see the peg and hole. To collaborate effectively, the robot should reason about the human's visual perspective and infer that the human is able to see the peg and hole (despite not knowing what he actually sees). It can then query the human for relevant information. 

Enabling robots to perform such perspective-taking is challenging, especially when the environment is partially-observable. Prior works on perspective-taking in Human-Robot Interaction (HRI) focus mainly on hand-crafted models (e.g., \cite{trafton2005enabling, berlin2006perspective, breazeal2006using, johnson2005perceptual, milliez2014framework}) or learning deterministic models in fully-observable environments~\cite{chen2021visual}. Unfortunately, handcrafted models do not easily scale to complex real-world environments with high-dimensional observations, and learnt deterministic models do not capture uncertainty in the beliefs or world state. 

In this paper, we take a step towards addressing this gap and propose a perspective-taking approach based on deep latent state-space world models. Given only partial observations (e.g., camera images), our \textbf{L}atent \textbf{E}mission-\textbf{A}ugmented \textbf{P}erspective-\textbf{T}aking (LEAPT) is able to infer  what another agent observes and knows. This is achieved by sampling a set of possible world states from the robot's underlying world or ``self'' model (Fig.\ref{fig:moti-example}.B.). Using these sampled states along with human pose information, we predict potential human observations, which are then used to infer human beliefs using a modified self-model~\cite{chen2022mirror}.

We find that the self-model requires careful design to be effective --- a standard latent state-space model with a single time-dependent latent variable did not properly capture uncertainty in the world state (resulting in samples that were blurry ``averages'' of training observations). To address this issue, we impose an additional model structure by decomposing the latent state into two variables that represent (i) information that the robot can fully observe ($s_t^R$) and (ii) information that is not revealed by the robot's observations ($h_t^R$). By conditioning $h_t^R$ on $s_t^R$, we explicitly train the model to generate possible task states given only partial information. Moreover, by minimizing the Kullback-Leibler (KL) divergence between task-complete observations (available during training) and the robot's estimation in latent space, the model learns to properly estimate uncertainty about the state. 

Experiments on various tasks in a simulated domain show that our model can better estimate the visual perceptions and beliefs of other agents compared to a deterministic approach and a standard latent-state space model. Specifically, we show our model enables robots to model false beliefs and better estimate another agent's (and it's own) beliefs during interaction and communication. We also report on a human-subject study using Virtual Reality (VR), which shows that LEAPT is better at predicting the human's belief about the world.

In summary, this paper makes the following contributions:
\begin{itemize}
    \item LEAPT --- a perspective-taking approach based on a latent state-space model --- for partially-observable environments; 
    \item A decomposed model structure and corresponding ELBO training loss that influences the model to better capture uncertainty over possible world states; 
    \item Empirical results that show that LEAPT outperforms competing methods and ablated variants.
\end{itemize}
\section{Background \& Related Work}
\label{sec:related_work}

Perspective-taking \cite{newcombe1992children, pearson2013review, ryskin2015perspective} is considered essential for many useful social skills in human-human interaction, such as understanding social dynamics and relationships. Within the context of robotics, perspective-taking can also occur during human-robot interaction 
(HRI)~\cite{trafton2005enabling, zhao2022spontaneous}. In the following, we briefly mention prior works on enabling artificial agents to perform visuospatial persepective taking and mentalizing (reasoning about another agent's mental states).

\para{Cognitive Architectures.} To enable efficient HRI, earlier works proposed cognitive architectures using hand-crafted rules and perspective-taking models~\cite{trafton2005enabling, berlin2006perspective, breazeal2006using, johnson2005perceptual, milliez2014framework}. Although these models can be effective, they are generally time-consuming to create and do not lend themselves easily to scenarios with high-dimensional observations (e.g., images). Potentially, modern physics simulators \cite{li2023behavior, gong2023arnold, chao2022handoversim} could be embedded within the cognitive architecture of the robot to perform perspective-taking. However, uncertainty estimation within such a setup remains computationally expensive --- one would need to either simulate multiple possible observations and track beliefs over time. 

\para{ML-based Models.} An alternative approach is to leverage machine learning (ML) to predict other agents' perspectives~\cite{chen2021visual, labash2020perspective} and apply planning~\cite{chen2021visual} (or reinforcement learning~\cite{labash2020perspective}) to generate robot behaviors. For example, in a hide-and-seek task~\cite{chen2021visual}, a robot hider predicts what the seeker sees using deterministic deep neural networks and then applies planning to avoid the seeker. Recent work learned a Q function conditioned on both  ego-agent and other agents observations in a dominant-and-subordinate-monkey task~\cite{labash2020perspective}. Unlike LEAPT, these works use deterministic models or assume access to other agents' observations. 

\para{Agent Modeling and Communication in Multi-Agent Systems} Our work is also related to opponent modeling in a multi-agent system. In this line of research, one approach is to optimize policies using a latent embedding that encodes the (predicted) trajectories of other agents~\cite{papoudakis2020variational, papoudakis2021agent, rabinowitz2018machine}. These methods either employ deterministic approaches when deployed \cite{papoudakis2021agent, rabinowitz2018machine} or model uncertainty in the observation space~\cite{papoudakis2020variational}. In the field of human-robot communication, recent works modify generated observations and track human belief using learned models~\cite{chen2022mirror, reddy2021assisted}. Unlike LEAPT, these methods assume that the robot is able to fully observe the environment at test time. LEAPT is designed to operate in \emph{partially-observable environments} for tasks where visual and conceptual (belief) perspective-taking is helpful.
\section{Latent Emission-Augmented Perspective-Taking (LEAPT)}

\para{Objective.} We aim to to enable a robot to perform:
\begin{itemize}
    \item \emph{Visual perspective-taking}: estimate what another agent observes at a given time;
    \item \emph{Conceptual perspective-taking}: Infer what the agent believes given the observations they have made so far, which is a form of a set of latent states.
\end{itemize}
We assume that the robot has access to its own on-board sensors and knows the other agent's pose. In addition, during training, the robot has access to a world-view that can provide task-complete information; this information is \emph{not} available during test time.

In this section, we describe our main contribution: the Latent Emission-Augmented Perspective-Taking (LEAPT) model. At its core, LEAPT uses a multi-modal latent state-space model (MSSM)~\cite{Chen2021MuMMI,chen2022mirror} that is modified by decomposing the latent state variable. 
We begin with a discussion of the standard MSSM and its limitations, which we then address by factorizing the latent state representation. This enables the robot to (i) better track its own uncertainty over the world state, and (ii) track the beliefs of other agents. We then illustrate how this model can be used for both visual and conceptual (belief) perspective taking via sampling.

\begin{figure} [!t]
    \centering
    \includegraphics[width=1.0\columnwidth]{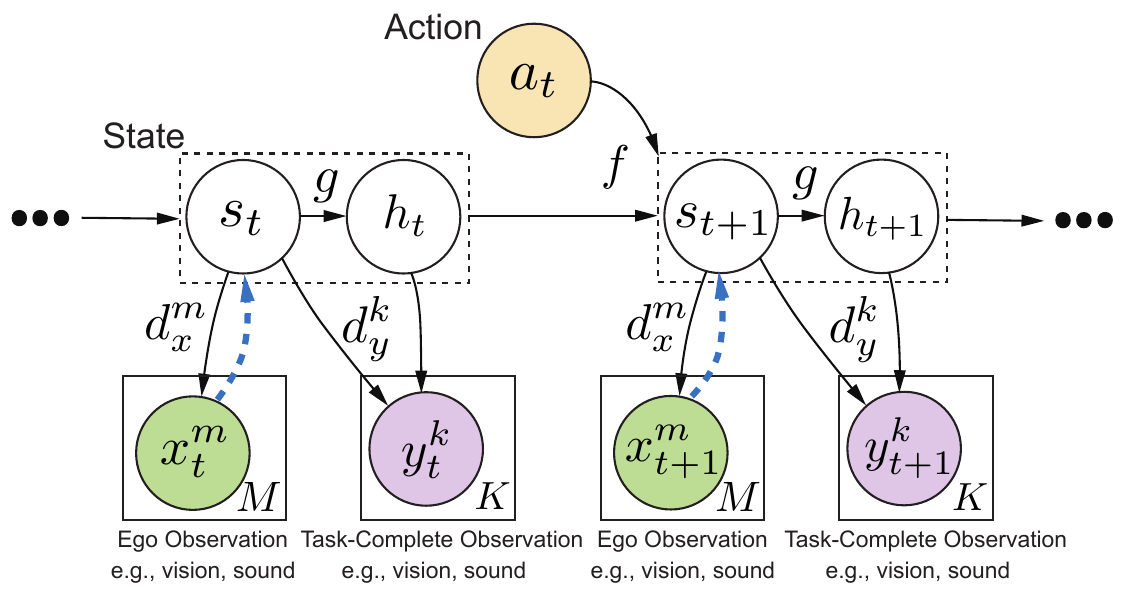}
    \caption{\small The LEAPT Decomposed Latent State-Space Model. Circle nodes represent random variables. The latent variables are decomposed into two parts, $z_t = [h_t, s_t]$. 
    Inference networks $q_\phi$ are shown using blue dotted line arrows; we see that $q(s_t|x_t^{1:M})$ computes $s_t$ (the part of the world that robot can fully observe), whilst $p(h_{1}|s_{1})$ and $p(h_{t}|s_{t-1, t}, h_{t-1})$ generates the part that robot cannot observe.
    }
    \label{fig:pgms}
    \vspace{-1em}
\end{figure}

\subsection{LEAPT Latent State-Space Model}
\label{sec:MSSM}

\para{Standard MSSM World Model.} Our robot maintains a world model --- called the \emph{self-model} --- on which it can perform inference and planning. We build upon the MSSM~\cite{Chen2021MuMMI,chen2022mirror}, which captures how the environment changes over time given actions taken by the robot. In general, the robot is unable to directly perceive the entire world state and we distinguish between two kinds of observations: 
\begin{itemize}
    \item Ego observations $x^m_t$  at time $t$ for $m=1, \dots, M$ sensory modalities, which are the agent's observations from its own perspective (e.g., a camera mounted on the robot's arm) and can be accessed in both training and testing;
    \item Task-Complete observations $y^k_t$ at time $t$ for $k=1,\dots,K$ modalities, which contains all sufficient information about world state to complete the task and \emph{can only be accessed during training}.
\end{itemize}

Similar to standard state-space models, the MSSM assumes Markovian transitions that are only dependent on the current state and the action $a_t$ taken by the agent. 

\para{Limitations of the standard MSSM.} One can train the MSSM using variational inference by optimizing the evidence lower bound (ELBO):
\begin{align}
\label{eqn:elbo}
     \mathcal{L}_e = & \sum_{t=1}^{T}\displaystyle  \mathop{\mathbb{E}}_{q_{\phi}(z_{t})}\left[\sum_{m=1}^M\log p_{\theta}(x^m_{t}|z_{t}) + \sum_{k=1}^K\log p_{\theta}(y^k_{t}|z_{t})\right]\nonumber\\  
    & - \sum_{t=2}^{T}\displaystyle\mathop{\mathbb{E}}_{q_{\phi}(z_{t-1})}\left[\DKL\left[q_{\phi}(z_{t}) \| p_{\theta}(z_{t}|z_{t-1},a_{t-1})\right]\right]\nonumber\\
    & -\displaystyle\mathop{\mathbb{E}}_{q_{\phi}(z_1)}\left[\DKL\left[q_{\phi}(z_{1}) \| p_{\theta}(z_1)\right]\right]
\end{align}
where the variational distribution $q_{\phi}(z_t)$ is shorthand for $q(z_t|u_{\phi}(x_{1:t}^{1:M}, y_{1:t}^{1:K}))$. This variational distribution can be modeled using an inference network $u_{\phi}$ that takes as input observations (\emph{both} ego observations $x_{1:t}^{1:M}$ and task-complete observations $y_{1:t}^{1:K}$) and outputs a distribution over $z_t$. In theory, once trained, $q_{\phi}(z_t)$ can be used to sample state $z_t$ given only agent's \emph{ego} observations $x^{1:M}_{1:t}$, i.e., the $z_t$'s will be used to enable perspective taking. 

The problem lies in using this inference network during test time when the task-complete observations are missing. One simple solution is to randomly drop the task-complete observations during training~\cite{Chen2021MuMMI, chen2022mirror, wu2018multimodal} by zero-ing out $y_{1:t}^{1:M}$. Unfortunately, our preliminary experiments results showed that this approach led to poor variational distributions: $q_\phi(z_t)$ was `narrow' (over-confident) when $x_{1:t}^{1:M}$ did not contain sufficient information of the world state. The image samples from $q_\phi(z_t)$ tended to be blurry ``averages'' of the data the model was trained with (see Fig \ref{fig:fetch-tool-examples} for examples). 

One possible reason for this undesirable behavior is that the MSSM is trained to encode $y^{1:K}_{t}$ into $z_t$ via the reconstruction term $\mathop{\mathbb{E}}_{q_{\phi}(z_{t})}\left[\log p_{\theta}(y^k_{t}|z_{t})\right]$. 

If we drop $y_{1:t}^{1:K}$, $q(z_t|u_{\phi}(x_{1:t}^{1:M}))$ may lack sufficient information to generate a sample representative of the world state. In such cases, the model learns to ``hedge'' by generating (decoding) the mean observation, which corresponds to a narrow $q_{\phi}(z_{t})$ centered around a specific latent state. 

\para{Decomposed Latent State.}
As a remedy, we propose to restructure the MSSM to have a more explicit generation process. This modified structure is illustrated in  Fig.\ref{fig:pgms}. Intuitively, we would like the model to learn how to generate task-complete observations \emph{only} using ego observations. As such, we decompose the latent state into two disjoint parts, $z_t = [s_t, h_t]$, where $s_t$ represents the observable part of the world and $h_t$ represents information that the robot cannot observe directly. By conditioning $h_t$ on $s_t$ in $p(h_{t}| g_{\theta}([s, h ,a]_{t-1},s_{t}))$, the model explicitly represents our desired computation using a neural network $g_{\theta}$. 

Given this decomposed latent structure, the joint distribution factorizes as:
\begin{align}
p_\theta(&x^{1:M}_{1:T},y^{1:K}_{1:T},s_{1:T},h_{1:T},a_{1:T-1}) =  \nonumber\\
& p_\theta(h_{1}|s_{1})p(s_1)\prod_{t=2}^{T}\left[\prod_{m=1}^{M}p_\theta(x^m_{t}|s_{t})\right] \left[\prod_{k=1}^{K}p_\theta(y^k_{t}|s_{t}, h_{t})\right] \nonumber\\
    & p_\theta(s_{t}|s_{t-1},h_{t-1}, a_{t-1})p_\theta(h_{t}|s_{t-1, t},h_{t-1}, a_{t-1})
\end{align}
where each of the factorized distributions is modeled using deep neural networks parameterized by $\theta$:
\begin{align}
    & \mathrm{\textrm{Transition:}} \ p_\theta([s,h]_{t}|[s,h,a]_{t-1}) \nonumber\\
    & \quad \quad = p(s_{t}| f_{\theta}([s,h, a]_{t-1}))p(h_{t}| g_{\theta}([s, h ,a]_{t-1},s_{t}))\\
    & \mathrm{\textrm{Ego Observations:} } \  p_\theta(x^m_{t}|s_{t}) = p(x^m_{t}| d^m_{x_\theta}(s_{t})) \\
    & \mathrm{\textrm{Task-complete Observations:} } \  p_\theta(y^k_{t}|[s, h]_{t}) \nonumber\\
    & \quad \quad = p(y^k_{t}| d^k_{y_\theta}([s, h]_{t}))
\end{align}

\para{Model Training.} Similar to the basic MSSM, we optimize the ELBO, but with two variational distributions $q(s_t|v_{\psi}(x_{1:t}^{1:M}))$ and $q(h_t|l_{\eta}(y_{1:t}^{1:K}))$ over the latent state variables $s_t$ and $h_t$. Note that $v_{\psi}$ and $l_{\eta}$ are two different inference networks. For notational simplicity, we will drop the explicit dependence on the conditioning variables and write the ELBO as,
\begin{align}
\label{eq:delbo}
     \mathcal{L}_e & =  \sum_{t=1}^{T}\Big(\displaystyle  \mathop{\mathbb{E}}_{q_{\psi}(s_{t})}\left[\sum_{m=1}^M\log p_{\theta}(x^m_{t}|s_{t})\right] \nonumber\\
     & + \displaystyle  \mathop{\mathbb{E}}_{q_{\psi}(s_{t})q_{\eta}(h_{t})}\left[\sum_{k=1}^K\log p_{\theta}(y^k_{t}|[s,h]_{t})\right]\nonumber\nonumber\Big)\\  
    & - \sum_{t=2}^{T}\Big(\displaystyle\mathop{\mathbb{E}}_{q_{\psi}(s_{t-1})q_{\eta}(h_{t-1})}\left[\DKL\left[q_{\psi}(s_{t}) \| p_{\theta}(s_{t}|[s,h, a]_{t-1})\right]\right]\nonumber\\
    & - \displaystyle\mathop{\mathbb{E}}_{q_{\psi}(s_{t-1})q_{\eta}(h_{t-1})}\left[\DKL\left[q_{\eta}(h_{t}) \| p_{\theta}(h_{t}|[s,h, a]_{t-1}, s_{t})\right]\right] \nonumber\Big)\\ 
    & - \displaystyle\mathop{\mathbb{E}}_{q_{\psi}(s_1)}\left[\DKL\left[q_{\eta}(h_{1}) \| p_{\theta}(h_{1}|s_1)\right]\right] \nonumber\\
    & -\DKL\left[q_{\psi}(s_{1}) \| p_{\theta}(s_1)\right]
\end{align}

In contrast to the MSSM, we don't need to discard $y^{1:K}_{1:t}$ during the training process. Instead, we substitute $q_{\eta}(h_t)$ with $p_{\theta}(h_t|[s,h, a]_{t-1}, s_{t})$ during the testing phase (or with $p_{\theta}(h_{1}|s_1)$ when $t=1$). This ELBO variant enables $q_{\eta}(h_t)$ to convert $y^{1:K}_{1:t}$ into a latent distribution, such as a Gaussian or Categorical distribution. By minimizing $\DKL\left[q_{\eta}(h_{t}) \| p_{\theta}(h_{t}|[s,h, a]_{t-1}, s_{t})\right]$ and $\DKL\left[q_{\eta}(h_{1}) \| p_{\theta}(h_{1}|s_1)\right]$, $p_{\theta}(h_{t}|[s,h, a]_{t-1}, s_{t})$ and $p_{\theta}(h_{1}|s_1)$ are trained to generate potential $h_t$ using observations $s_{1:t}$. In a nutshell, this latent state-space model decomposition captures task-complete uncertainty at the latent level. By carefully setting $q$ and $p$, e.g., Gaussian, we can calculate these KL divergence terms exactly. We found that this decomposition resolves the issue of suboptimal latent state estimations and reconstructions, enabling us to create a self-model that produces credible samples. Please refer to Fig. \ref{fig:fetch-tool-examples} and \ref{fig:human-belief-pca} for examples.
\subsection{Visual Perspective-Taking}
\label{sub:sec:vpt}
We use sampled task-complete observations $y^{1:K}_t$ to generate relevant observations from different perspectives. Specifically, we train a visual perspective-taking model $\hat{x}^{1:M}_{t} = d_\chi(y^{1:K}_{t}, \omega_t)$, parameterized by $\chi$, that produces observations given an agent's pose $\omega_t$ at time $t$ and $y^{1:K}_t$. 

Training can be performed with data gathered by the robot $\robot$ during roll-outs in the environment. Given trajectories of the form $\left\{(x^{1:M,\robot}_{t}, y^{1:K}_{t}, \omega_t^\robot)\right\}_{t=1}^{T}$ collected by the robot, we learn $\chi$ by minimizing the following loss:

\begin{align}\label{eqn:pt_learning}
   \argmin_\chi \ &\mathcal{L}(\chi) = \sum_{t=1}^T \sum_{m=1}^M\left(\hat{x}^{m}_{t} - x^{m}_{t}\right)^2 \nonumber
\end{align}

Once trained, we can use the decomposed latent state space model and $d_\chi$ to obtain samples of another agent's observations. The process is straightforward: given robot observations $x^{1:M,\robot}_{t}$ and the agent's pose $\omega_t^\human$, we sample a task-complete observation $y^{1:K}_{t}$ via 
\begin{enumerate}
    \item Sample the robot's belief state $s_t^\robot \sim q_\psi(s_t|(\hat{x}^{1:M, \robot}_{t}))$ and $h_t^\robot \sim p_{\theta}(h_{t}|[s^\robot,h^\robot, a]_{t-1}, s^\robot_{t})$ with $h_1^\robot \sim p_{\theta}(h_{1}|s^\robot_{1})$.
    \item Sample the task-complete observation $\hat{y}^{1:K}_{t} \sim p_{\theta}(y^{1:K}_{t}|s_t^\robot, h_t^\robot)$.
    \item Given human position, generate human observations $\hat{x}^{1:M, \human}_{t}= d_\chi(\hat{y}^{1:K}_{t}, \omega_t^\human)$
\end{enumerate}

As an aside, one may be curious why we did not directly predict $x^{m}_{t}$ using $h_t^\robot$ and $s_t^\robot$. Our preliminary experiments indicated that this direct approach caused the model to disregard the poses $\omega_t^\human$, resulting in poor samples. This behavior stems from training the perspective-taking model on tuples of \emph{robot} pose and \emph{robot} ego observations $x^{1:M,\robot}_{t}$. By including the robot's latent states $h_t^\robot$ and $s_t^\robot$ (derived from $x^{1:M,\robot}_{t}$) the model is able to make predictions directly from them, overshadowing the importance of the agent poses. As a result, substituting $\omega_t^\robot$ with human pose $\omega_t^\human$ during test time leads to the failure of the perspective-taking model in generating accurate observations from the other agent's perspective.
\subsection{Conceptual (Belief) Perspective-Taking} \label{sub:sec:bpt}
Next, we turn our attention to how the self-model can be used to estimate human belief, which is a form of conceptual perspective taking. We follow prior work~\cite{chen2022mirror} and use a variant of the robot's self-model. By coupling the human model and robot's self-model together as shown in Fig.\ref{fig:moti-example}.B., we sample belief states in the human model: 
\begin{enumerate}
    \item Generate the human observations $\hat{x}^{1:M, \human}_{1:t}$ by visual perspective-taking.
    \item Sample the human's belief state $s_t^\human \sim q_\psi(s_t|(\hat{x}^{1:M, \human}_{1:t}))$ and $h_t^\human \sim p_{\theta}(h_{t}|[s^\human,h^\human, a]_{t-1}, s^\human_{t})$ with $h_1^\human \sim p_{\theta}(h_{1}|s^\human_{1})$.
\end{enumerate}
Unlike existing approaches, LEAPT can capture a (sample-based) distribution over possible observations and beliefs. This enables the robot to perform epistemic reasoning, i.e., the robot can infer that the other agent is aware of some information even though the robot doesn't know exactly what that information is. We will see examples of this ability in the experiments.

\section{Simulated Human Experiments}
\label{sec:experiments}
\begin{figure*}
    \centering
    \includegraphics[width=0.85\textwidth]{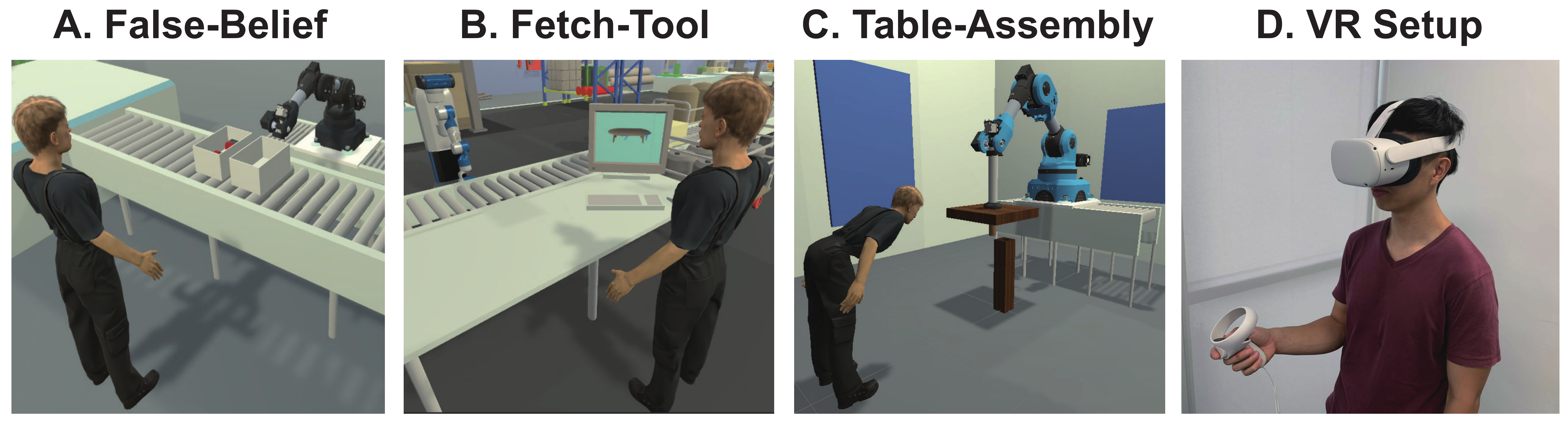}
    \caption{\small Experimental domains. (\textbf{A}) A False-Belief test, where the robot \emph{cannot} see what is inside the box but the human can. The human first observes the position of the drill (in the left or right box) and then walks away. Then, unobserved by the human, the robot will either switch the box or not. The robot's task is to infer whether the human would believe that the objects remain in their original position. (\textbf{B}) In the Fetch-Tool task, a human is sitting in front of his workstation and observes a randomly chosen object on his computer screen. There are six different types of objects, which are characterized by their color and type. The robot's task is to reason about human uncertainty and what he sees without observing the object, using only communication information from the human. However, communication of the object has three levels of ambiguity: type, colour and both the type and colour of the object. Once the robot figures out what the human sees, it can subsequently fetch the appropriate tool to assemble the object. (\textbf{C}) In the Table-Assembly, the robot's task is to fit the peg of a table piece into a hole with guidance from the human. However, the hole may be occluded depending on the agent's position. As the robot is holding on to the table, only the human is able to observe the relative distance between the peg and hole. The robot's task is to reason whether the human can see the hole so that the robot can move the table to match it. (\textbf{D}) Our VR setup in the human-subject experiments used an Occulus Quest headset.}
    \label{fig:exp_domains}
\end{figure*}

In the following section, we describe experiments with simulated humans on a False-Belief test, a Fetch-Tool task, and a Table-Assembly task (Fig. \ref{fig:exp_domains}.). 
Using simulated humans allowed us to compare the learned models against ground truth observations and beliefs. Experiments with real human subjects are described in the next section.

\subsection{Experimental Setup} 
\para{Domains.} Fig. \ref{fig:exp_domains}. summarizes the three domains used in our experiments. All domains involve two agents: a robot and a human. The observations of the environment (both ego and task-complete) are images. We assume the agents can communicate via specific speech symbols in the Fetch-Tool and Table-Assembly tasks (these are simply modeled as observations as in~\cite{chen2022mirror}). We assume the robot can always observe the human's pose.

\para{Simulated Humans.} We assume that the human is a Bayes-rational agent that updates their beliefs using ego-centric observations. For example, in the False-Belief test, the simulated human will believe that the drill remains in the same position when they walk away. Specifically, the simulated human's belief is a Bernoulli distribution representing where the drill is (left or right box). In the Fetch-Tool task, the simulated human belief is a Categorical distribution over which object they see on the screen (six classes). In the Table-Assembly task, we use a Gaussian distribution over the relative distance between the peg and hole.  

\para{Compared Methods.} In total, we compare four different
perspective-taking methods:
\begin{itemize}
    \item Baseline-S: This baseline uses the standard MSSM and is trained with the ELBO in Eq.~\ref{eqn:elbo}. We use bi-directional GRU as the main component of the inference network.
    \item Baseline-D: This baseline is similar to Baseline-S except that the latent space is deterministic.
    \item LEAPT-GRU: Our decomposed latent state space model with a GRU-based inference network. The model is trained with the ELBO in Eq. \ref{eq:delbo}.
    \item LEAPT-Transformer: This model is the same as the LEAPT-GRU, except we use a Transformer-based~\cite{vaswani2017trans} inference network. 
\end{itemize}
All the models are trained on the same data and epochs within each domain (False-Belief: 30 trajectories with length 5, 100 epochs; Fetch-Tool: 100 trajectories with length 5, 400 epochs; Table-Assembly: 300 trajectories with length 10, 700 epochs). The trajectories used for training are generated by taking random actions. At every time step $t$, the inference time is less than 0.05s for all models on a NVIDIA GeForce RTX 2080 Ti. Source code is available at \url{https://github.com/clear-nus//perspective-taking-dmssm}.

\para{Evaluation}
We first evaluate whether the robot maintains a correct belief about the world. Then, we evaluate the robot's ability to perform visual and belief perspective-taking, i.e., infer what the other agent sees and believes. As the direct evaluation of latent states is challenging, we utilize pre-trained classifiers or regression models on the samples $x$ and $y$ that are decoded from the latent states. Denote the predicted labels as $c(x)$ or $c(y)$. The exact predicted quantities was task-dependent: i) In the False-Belief test, we predict where the drill is; ii) In the Fetch-Tool task, we predict the object present on the screen (Fig. \ref{fig:exp_domains}.B.); iii) In the Table-Assembly task, we predict the relative position between human and table and the relative distance between peg and hole (Fig. \ref{fig:exp_domains}.C.). We then compare the distribution of these predictions to a specified ground-truth distribution (assuming a Bayes-rational agent) via KL divergence (lower is better).

\emph{Evaluation of robot's belief:} We first decode $\hat{y}^\robot$ from a set of the robot's latent states. Then, we apply classifiers/regression models on $\hat{y}^\robot$ to obtain predictions $c(\hat{y}_{t}^\robot)$. The quality of the robot's belief is measured through the KL divergence $\DKL\left[\hat{p}(c(\hat{y}_{t}^\robot)|x_{1:t}^\robot) \| p(c({y}_{t}^\robot)|x_{1:t}^\robot)\right]$ between the empirical distribution, $\hat{p}$, and the ground truth distribution $p$ (representing what the robot should know about world given our experiment parameters). 

\emph{Evaluation of  visual perspective-taking:} We generate human observations $\hat{x}_{t}^\human$ via our visual perspective-taking method (Sec. \ref{sub:sec:vpt}). Then, we predict labels $c(\hat{x}_{t}^\human)$ and compute the KL divergence $\DKL\left[\hat{p}(c(\hat{x}_{t}^\human)|x_{1:t}^\robot) \| p(c({x}_{t}^\human)|x_{1:t}^\robot)\right]$. The distribution $p$ varies according to the amount of information available, e.g., in the Fetch-Tool task, if the robot has yet to communicate with the human, $p(c({x}_{t}^\human)|x_{1:t}^\robot)$ is uniform over all possible objects. After receiving a message that the object is brown, the ground truth distribution shrinks to equal probabilities across brown objects.

\emph{Evaluation of belief perspective-taking:} We begin by sampling from the human belief distribution (via belief perspective-taking in Sec. \ref{sub:sec:bpt}). Then, we generate labels $c(y_t^\human)$ and compute the Conditional KL (Cond KL) measure via $\mathbb{E}_{x_{1:t}^\human}\left[\DKL\left[\hat{p}(c(\hat{y}_{t}^\human)|x_{1:t}^\human)\| p(c({y}_{t}^\human)|x_{1:t}^\human)\right]\right]$, where $x_{1:t}^\human$ is sampled using visual perspective-taking. 

\begin{figure}
    \centering
    \includegraphics[width=1.0\columnwidth]{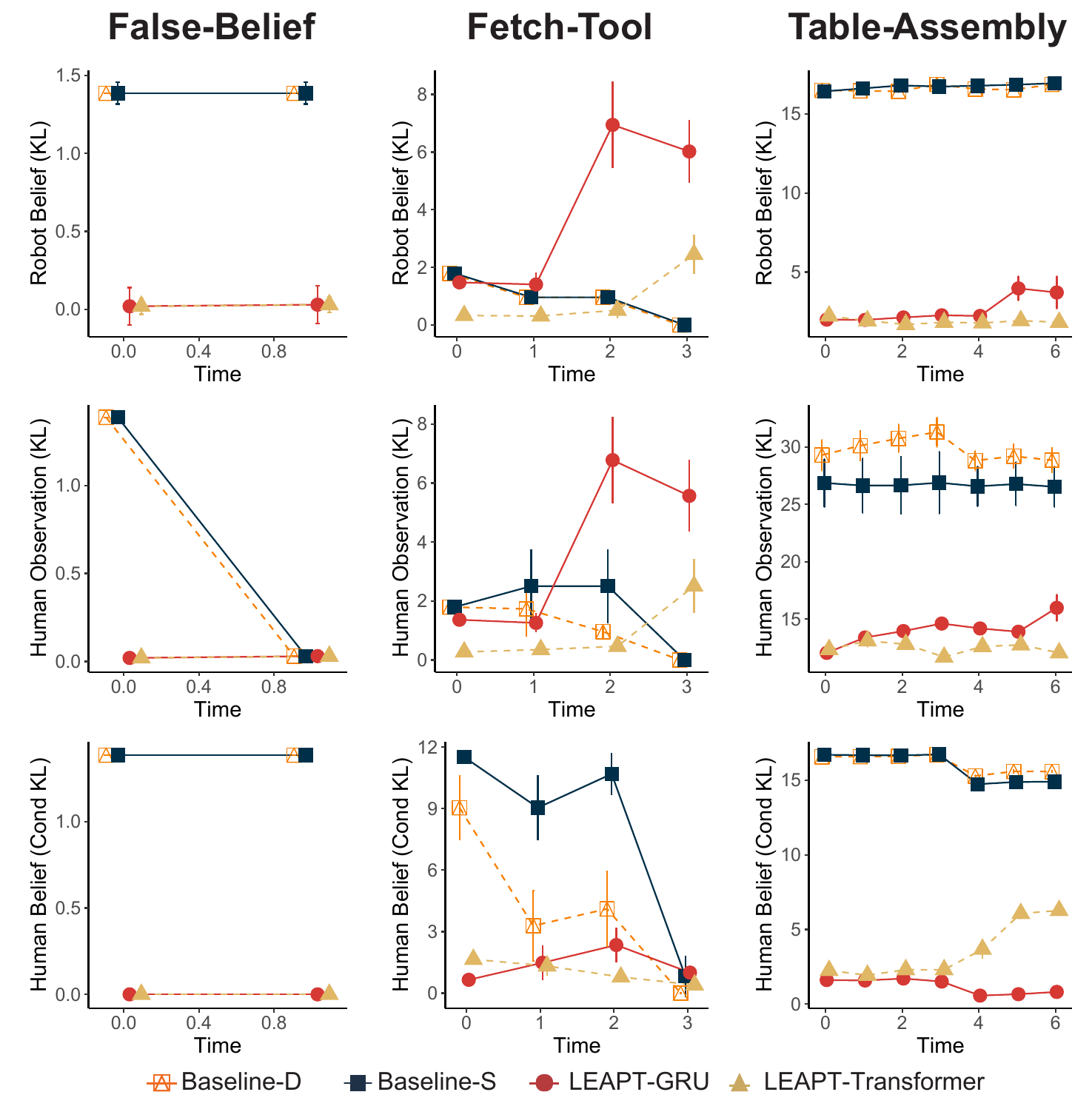}
    \caption{\small Evaluation of (visual \& conceptual) perspective-taking in three tasks using KL divergence as a measure (lower is better). In most cases, LEAPT-Transformer predicts the (simulated) human's observation and belief more accurately compared to the other methods.} 
    \label{fig:sim-obs_pi_KL_sim}
\end{figure}

\begin{figure}
    \centering
    \includegraphics[width=0.95\columnwidth]{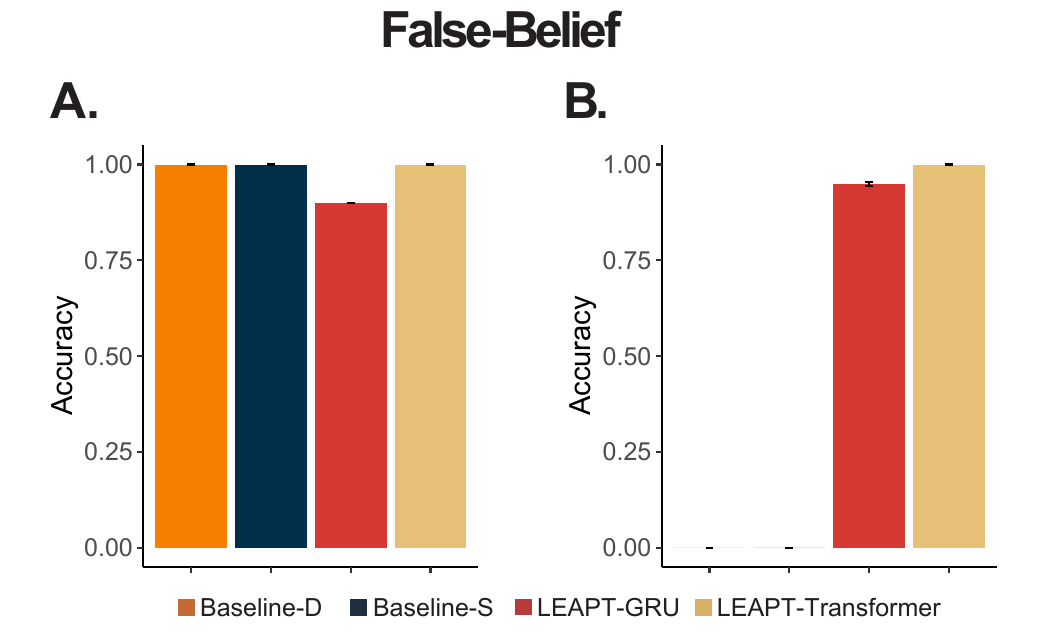}
    \caption{\small Accuracy of human belief prediction on the False-Belief Test. (A) boxes not switched by robot (B) boxes switched by robot.}
    \label{fig:sally-anne-accuracy}
\end{figure}

\subsection{Results and Analysis}
Due to space considerations, we highlight key results from our experiments. Additional plots are available in our online appendix: \url{https://github.com/clear-nus//perspective-taking-dmssm.}.

\para{Overall, LEAPT-Transformer maintains more accurate beliefs about the world compared to the baselines when given only ego observations.} We see in the first row of Fig. \ref{fig:sim-obs_pi_KL_sim} that the KL divergence tends to be smallest for the LEAPT variants in the False-Belief and Table-Assembly tasks. The sampled images from the baselines tended to be blurry, leading to poor identification of task-relevant features. For Fetch-Tool, the LEAPT-Transformer achieved better scores in the initial time-steps, but the baselines performed well when all needed information was available (at time-step 3). The LEAPT-Transformer state distribution was overly broad at the last time-step, but it generated the best samples overall over the course of the interaction. 

\begin{figure}
    \centering
    \includegraphics[width=0.99\columnwidth, keepaspectratio=true]{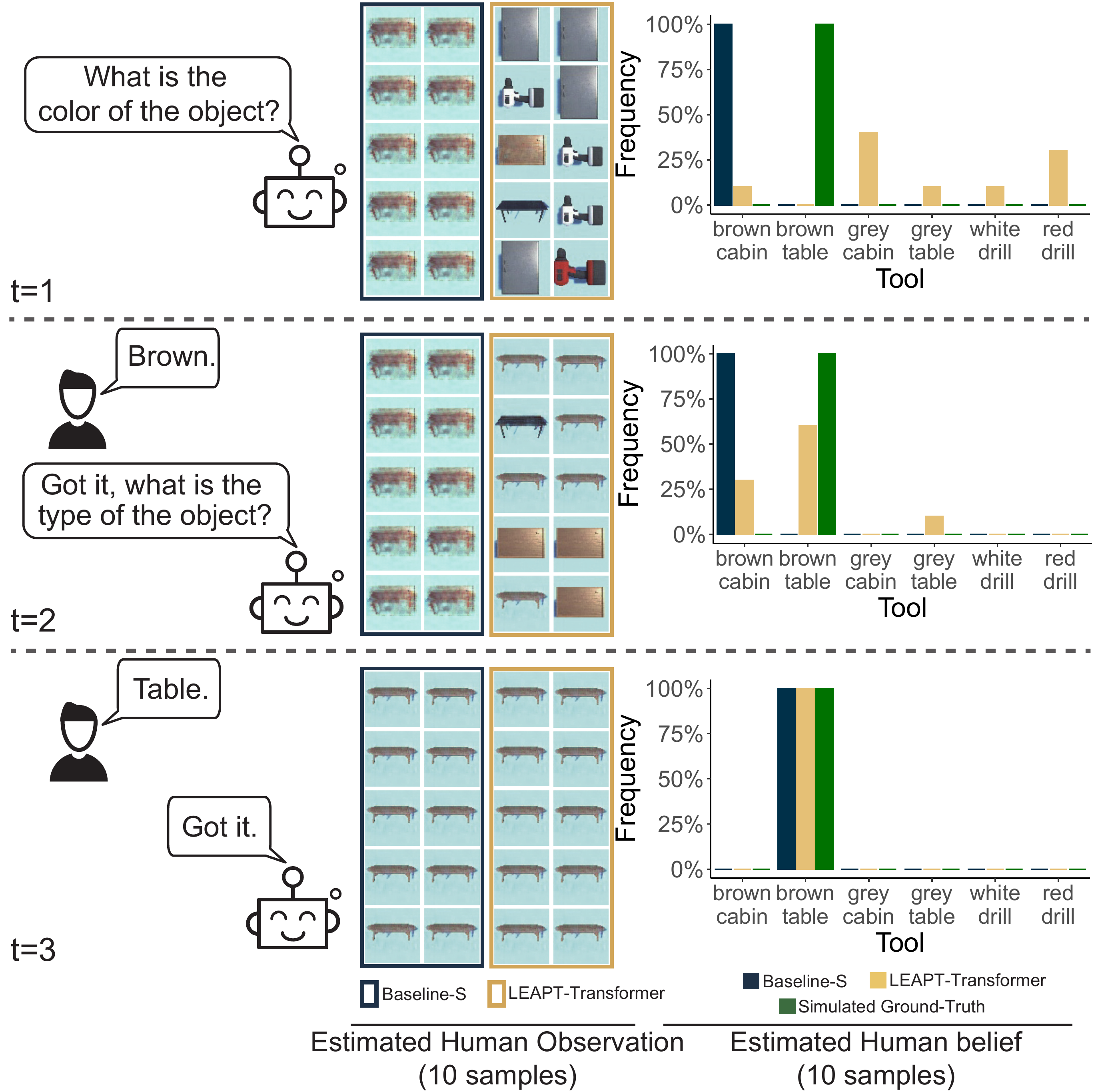}
    \caption{\small Fetch-Tool qualitative results: Baseline-S vs. LEAPT-Transformer. For each model, we sample 10 human observations and 10 human latent states (we decode latent states into image observations and apply a pre-trained classifier to obtain the estimated human beliefs). The simulated ground truth always shows the correct object. Compared to the baseline, the LEAPT model belief distributions are consistent with the information available at each stage of the communication.}
    \label{fig:fetch-tool-examples}
\end{figure}

\para{The LEAPT-Transformer achieves better visual perspective-taking given the robot's beliefs.} The second row of Fig. \ref{fig:sim-obs_pi_KL_sim} shows better scores by the LEAPT-Transformer, except (again) for the last time step of the Fetch-Tool task.
In general, the quality of the generated samples (both diversity and visual clarity) was positively correlated with better robot beliefs\footnote{For the False-Belief task, note the human doesn't return to the boxes so the generated observations are of a different area.}. The baselines tended to produce poor quality images, with sharp images only appearing when all task information was provided (See Fig. \ref{fig:fetch-tool-examples} for examples).

\para{The LEAPT models better infer the other agent's beliefs compared to the baselines.} This can be seen in the last row of Fig. \ref{fig:sim-obs_pi_KL_sim}, where the KL scores are overall better than the baselines. For the False-Belief test, a more interpretable result is shown in Fig. \ref{fig:sally-anne-accuracy} where we ask the robot to classify where the human thinks the object to be. When the boxes are switched, the baselines fail to capture that the human holds a false belief. 

\begin{figure}
    \centering
    \includegraphics[width=0.99\linewidth]{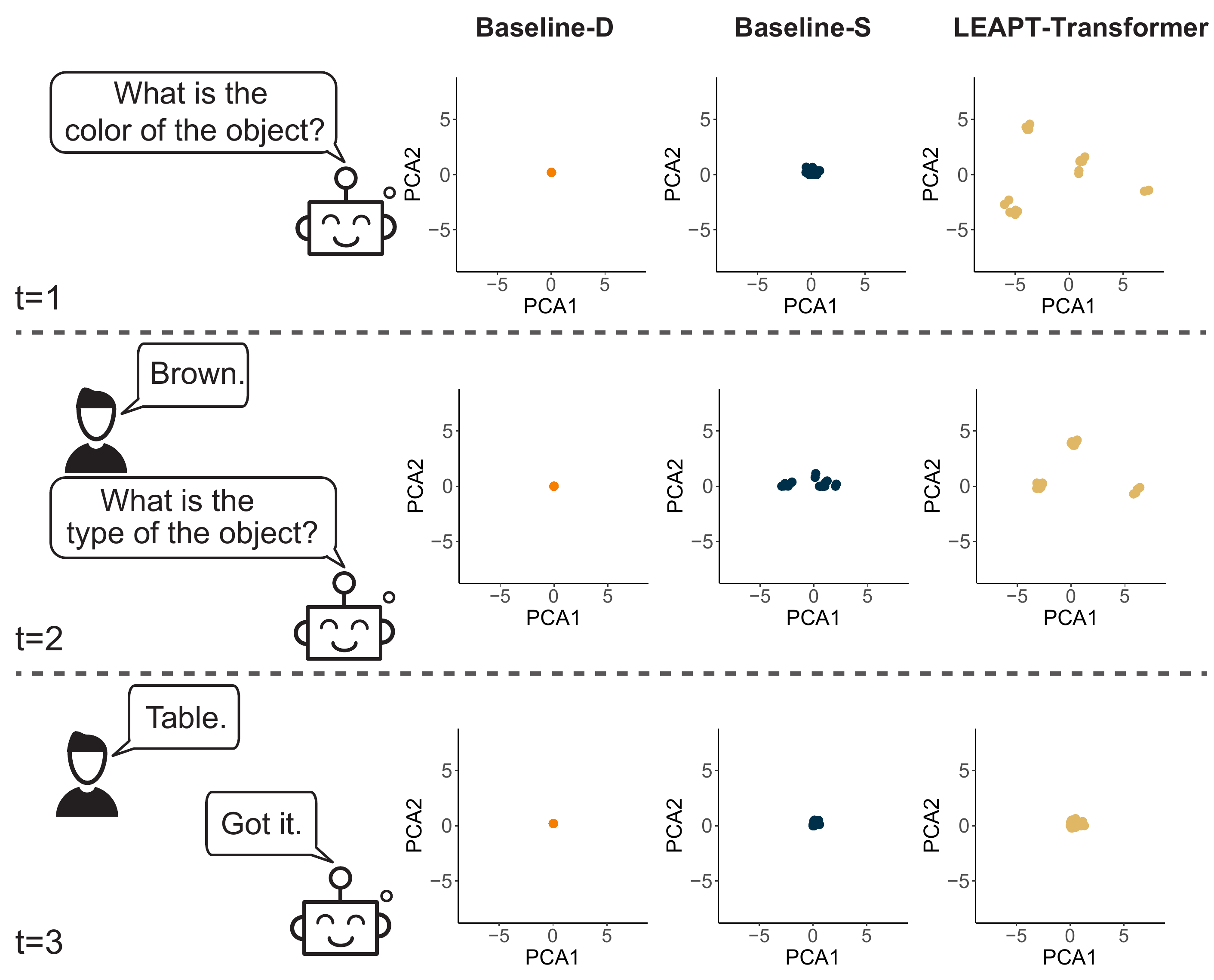}
    \caption{\small The visualization of estimated human belief (latent states) in the fetch tool 
    task (dimension reduced to two by PCA). Unlike the baselines, the LEAPT-Transformer maintains a reasonable distribution over possible objects when provided with communication.}
    \label{fig:human-belief-pca}
\end{figure}

For the fetch tool task, a 2D Principal Components Analysis (PCA) plot (Fig.\ref{fig:human-belief-pca}) suggest that the baseline models do not learn an appropriate latent space. It incorrectly maps different partial observations to the same cluster of latent points. On the other hand, LEAPT-Transformer correctly captures the distribution of latent states: with more communication coming from the human, the states converge. 

Fig.\ref{fig:fetch-tool-examples} illustrates the ability of the LEAPT-Transformer to predict human belief given partial information. At the start of the interaction, without any information, the LEAPT-Transformer  predicts a broad distribution over possible human observations and beliefs over all objects. After receiving communication that the object's color is brown, the uncertainty over human observations and human belief shrinks to the brown cabinet and table. Thereafter, it receives communication that the object type is a table, allowing it to determine the object in question and the distributions shrink to what the human actually sees. In contrast, the Baseline-S model consistently outputs a single image (Fig.\ref{fig:fetch-tool-examples}) which is often blurry, failing to correctly model uncertainty under partial observations (similar results were obtained for Baseline-D).

\begin{figure}
    \centering
    \includegraphics[width=0.90\columnwidth]{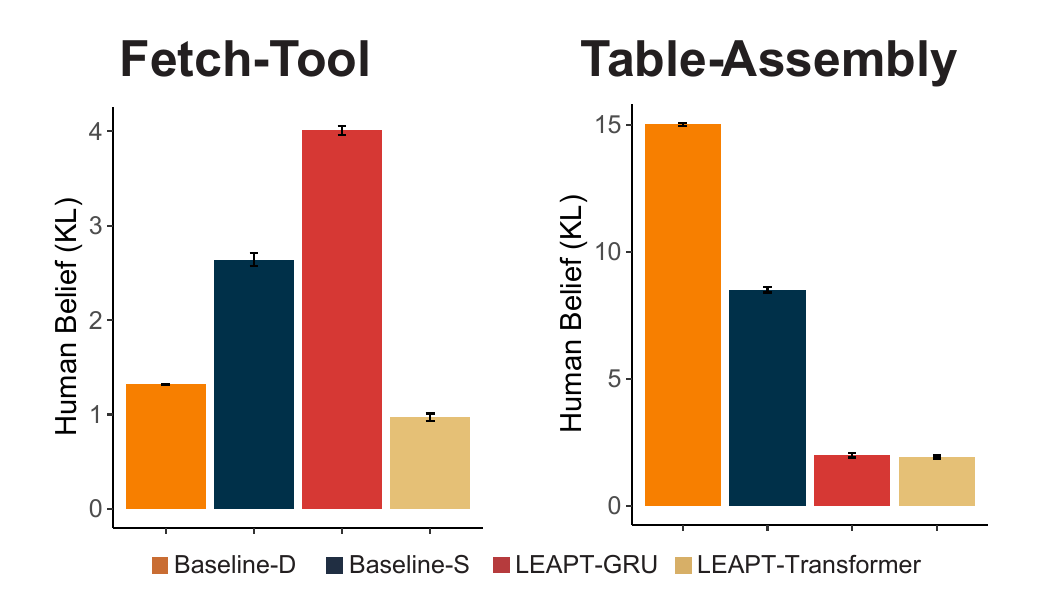}
    \caption{\small Human-subject experiment results: KL divergence between predicted and actual human belief. The scores were averaged over the episode length. Overall, the LEAPT-Transformer achieves the best performance (lowest KL divergence).}
    \label{fig:real-obs_pi_KL}
\end{figure}

\section{Human-Subject Experiments}

In this section, we describe preliminary human-subject experiments designed to test the viability of LEAPT when interacting with real humans. 
A total of 12 participants (mean age = 22.2,
3 females, 9 males) were recruited from the university community. The participants were asked to perform the Fetch-Tool and the Table-Assembly tasks using a VR setup (Fig.\ref{fig:exp_domains}.D). In the Fetch-Tool task, the participants sat in front of the (simulated) workstation, and the robot was situated on the other side. The robot queried about the type or color of the item shown on the screen, and the participants would respond accordingly. In the Table-Assembly task, the participants stood in front of the table, where they could see the peg and hole. The robot was on the other side, gripping the table, which visually occluded the peg and hole. The robot queried the participants on the relative distance of the hole from the peg and moved the table accordingly until the participants confirmed that the peg was aligned with the hole.

\para{Evaluation} Our evaluations are similar to the simulated human experiments described in the previous section. From the robot observations and human communication collected during the human-subject experiments, we use our models to derive the estimated empirical human belief. We then evaluate it with the ground truth distribution set based on the information provided in human communication. For instance, in the Fetch-Tool task, when a human communicated that they saw a table, the ground truth distribution was set as uniform over all kinds of tables. We measure the KL divergence $\DKL\left[p(c(\hat{y}_{t}^\human)|x_{1:t}^\robot)\|p(c({y}_{t}^\human)|x_{1:t}^\robot)\right]$, where $c$ is a classifier/regression model. 
 
\para{Results.} 
Overall, the LEAPT-Transformer attained the best performance. 
We compared the methods using repeated measures one-way ANOVA, followed by selected pairwise t-tests with adjusted $\alpha=0.008$ using Bonferroni correction. Fig.\ref{fig:real-obs_pi_KL}. shows that LEAPT-Transformer outperforms the baselines significantly in the Fetch-Tool task ($F_{3,33} = 966.324, p < 0.001$; LEAPT-Transformer vs. Baseline-D: $t(11) = 9.209, p < 0.001$ and LEAPT-Transformer vs. Baseline-S: $t(11) = 19.78, p < 0.001$) and the Table-Assembly task ($F_{3,33} = 7098.428, p < 0.001$; LEAPT-Transformer vs. Baseline-D: $t(11) = 140.874, p < 0.001$ and LEAPT-Transformer vs. Baseline-S: $t(11) = 52.906, p < 0.001$). 
\section{Conclusions}
This paper introduces LEAPT, a framework that empowers robots to conduct visual and conceptual perspective-taking in partially-observable environments. The key innovation is a specially-designed latent multi-modal state-space model, which enables more consistent beliefs to be maintained when observations are limited. Our experiments on three tasks show that LEAPT leads to better performance compared to deterministic approaches and standard latent-state space models when estimating the visual perceptions and beliefs of other agents.

Moving forward, we plan to apply LEAPT to other robot tasks and in real-world environments. In addition, LEAPT currently exhibits several limitations that would make for interesting future work. First, LEAPT requires human poses during testing. To address this, an alternative approach could involve an additional distribution over the poses given partial observations of the human and a learnt dynamics model.  Second, the model needs to be trained with sufficient views of the world to be effective. Inadequate training data may impede the model's ability to handle complex environments. Third, the human model in LEAPT is based on the robot self-model, thus, falls short in capturing certain aspects of the human, such as trust~\cite{chen2018planning, Lee_Fong_Kok_Soh_2020,kok2020trust} and emotions~\cite{ong2019applying}. Lastly, communication in LEAPT is based on simple, pre-defined sentence templates and we can integrate large language models (e.g., \cite{bowen23large, yaqi23translating}) to enhance LEAPT's communication ability.   

\balance 
\bibliographystyle{IEEEtran}
\bibliography{references}

\clearpage
\section{Appendix} \label{sec:AppendixAdditionalResults}

\subsection{Experiment Observations}

\begin{figure}[!htbp]
    \centering
    \includegraphics[width=0.5\linewidth, keepaspectratio=True]{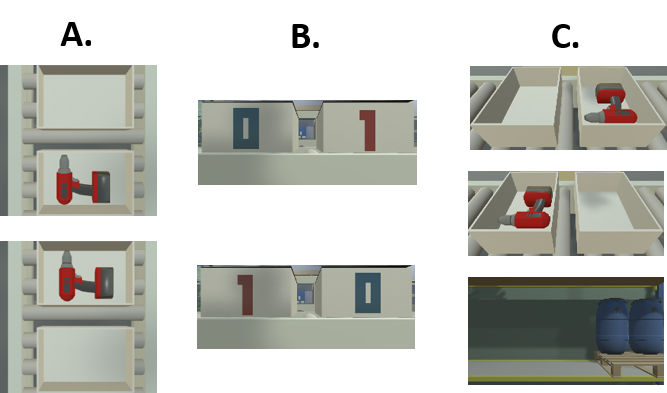}
    \caption{\small Sally-Anne environment observations. (\textbf{A}) Full observations. Drill at the bottom denotes that left of the robot and vice versa. (\textbf{B}) Robot partial observation. It can only observe if the boxes are switched or not. (\textbf{C}) Human partial observations. Human can see the location of the item when he is in front of the boxes.}
    \label{fig:sa_env_imgs}
\end{figure}

\begin{figure}[!htbp]
    \centering
    \includegraphics[width=0.5\linewidth, keepaspectratio=True]{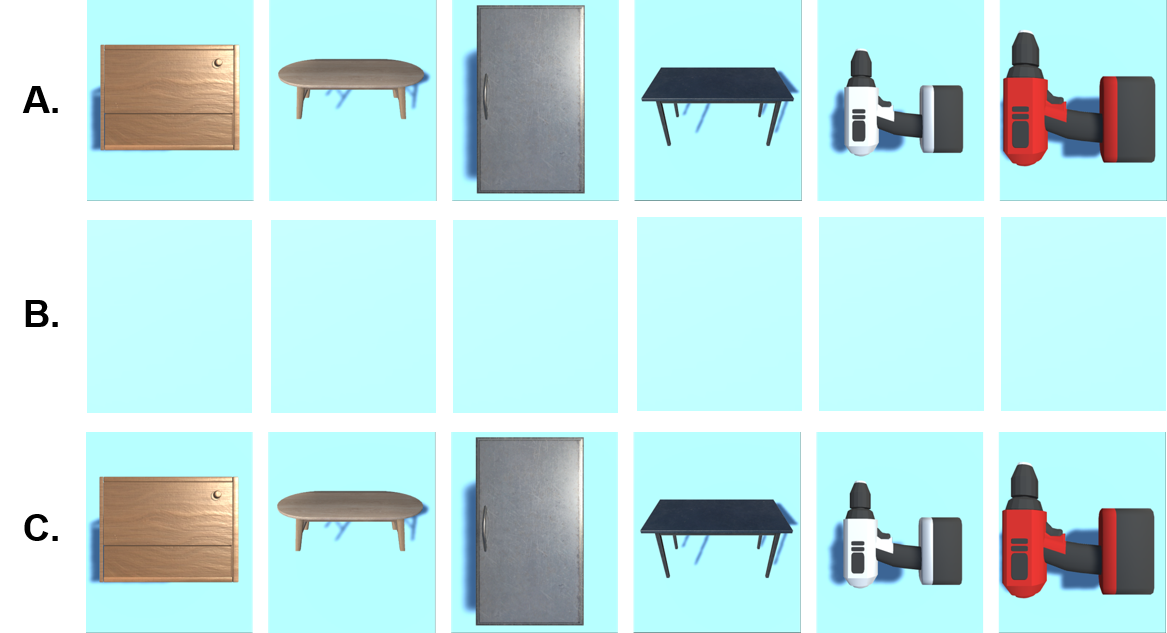}
    \caption{\small Fetch tool environment full observations. (\textbf{A}) Full observations. (\textbf{B}) Robot observations, the robot is looking at the table and does not observe the object. (\textbf{C}) Human observations.}
    \label{fig:ft_env_imgs}
\end{figure}

\begin{figure}[!htbp]
    \centering
    \includegraphics[width=0.45\linewidth, keepaspectratio=True]{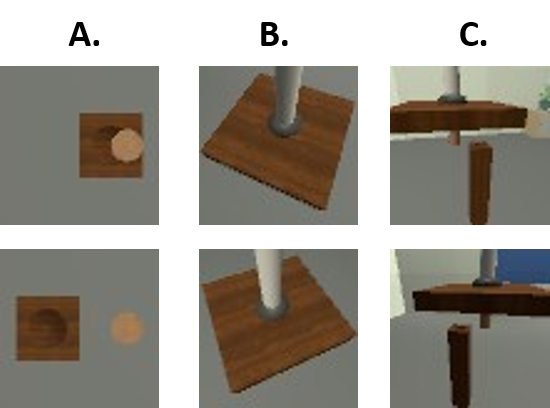}
    \caption{\small Table assembly environment observations. (\textbf{A}) Full observation which shows the top-down view of the position of hole (square) and table (circle). (\textbf{B}) Robot's partial observation. (\textbf{C}) Human partial observations.}
    \label{fig:ta_env_imgs}
\end{figure}

\end{document}